\documentclass[runningheads]{llncs}
\titlerunning{Machine Learning-Based Generalized Model for Finite Element Analysis}
\usepackage{graphicx}
\usepackage{caption}
\usepackage[margin=0.95in]{geometry}
\usepackage{subcaption}
\usepackage{tikz}
\usetikzlibrary{spy,calc}
\usepackage{hyperref}
\usepackage{verbatim}
\usepackage{booktabs,array} 
\usepackage{url}
\usepackage{xfrac}

\usepackage{siunitx}
\usepackage{microtype}
\usepackage{cite}
\usepackage{breqn}
\usepackage{amsmath,bm}

\usepackage[ruled,vlined]{algorithm2e}

\usepackage{pgfplots}
\pgfplotsset{compat=newest, legend style={at={(1,0.05)},anchor=south east}
}

\hypersetup{hidelinks,
backref=true,
pagebackref=true,
hyperindex=true,
breaklinks=true,
colorlinks=true,
urlcolor=blue,
bookmarks=true}

\usepackage{cleveref}

\usepackage{subcaption}

\crefname{section}{Sec.}{Sections}
\crefname{figure}{Fig.}{Figure}
\crefname{table}{Tab.}{Table}
\crefname{equation}{Equ.}{Equation}

\newcolumntype{L}[1]{>{\raggedright\arraybackslash}p{#1}}

\makeatletter
\DeclareRobustCommand\onedot{\futurelet\@let@token\@onedot}
\def\@onedot{\ifx\@let@token.\else.\null\fi\xspace}
\makeatother

\newcommand{\cf}{cf\onedot}
\newcommand{\ie}{i.\,e.,\xspace}

\makeatletter
\newcommand{\thickhline}{%
    \noalign {\ifnum 0=`}\fi \hrule height 1pt
    \futurelet \reserved@a \@xhline
}

\newcommand{\StressLFEM}{\textsc{Stress316L}\xspace}

\begin{document}

\title{Machine Learning-Based Generalized Model for Finite Element Analysis of Roll Deflection During the Austenitic Stainless Steel 316L Strip Rolling}

\author{Mahshad Lotfinia\thanks{Corresponding author.}\inst{1, 2}
\and Soroosh Tayebi Arasteh\inst{3, 4}}

\institute{Department of Theoretical Physics, University of Regensburg, Regensburg, Germany \and Department of Materials Science and Engineering, Sharif University of Technology, Tehran, Iran \and
Pattern Recognition Lab, Friedrich-Alexander-Universit\"at Erlangen-N\"urnberg, Erlangen, Germany \and
Harvard Medical School, Harvard University, Boston, MA, USA\\
\email{mahshad.lotfinia@alum.sharif.edu}}

\maketitle

\begin{abstract}

During the strip rolling process, considerable amount of the forces of the material pressure cause elastic deformation on the work-roll, \ie the deflection process. 
The uncontrollable amount of the work-roll deflection leads to the high deviations in the permissible thickness of the plate along its width. 
Therefore, the work-roll deflection should be controlled.
In the context of Austenitic Stainless Steels (ASS), due to the instability of the Austenite phase in a cold temperature, cold deformation leads to the production of Strain-Induced Martensite (SIM), which improves the mechanical properties. 
It leads to the hardening of the ASS 316L during the cold deformation, which causes the Strain-Stress curve of the ASS 316L to behave non-linearly, which distinguishes it from other categories of steels.
To account for this phenomenon, we propose to utilize a Machine Learning (ML) method to predict more accurately the flow stress of the ASS 316L during the cold rolling.
Furthermore, we conduct various mechanical tensile tests in order to obtain the required dataset, \StressLFEM, for training the neural network.
Moreover, instead of using a constant value of flow stress during the multi-pass rolling process, we use a Finite Difference (FD) formulation of the equilibrium equation in order to account for dynamic behavior of the flow stress, which leads to the estimation of the mean pressure, which the strip enforces to the rolls during deformation.
Finally, using the Finite Element Analysis (FEA), the deflection of the work-roll tools will be calculated.
As a result, we end up with a generalized model for the calculation of the roll deflection, specific to the ASS 316L.
To the best of our knowledge, this is the first model for ASS 316L which considers dynamic flow stress and SIM of the rolled plate, using FEM and an ML approach, which could contribute to the better design of the tolls.
Additionally, we compare our proposed numerical method against a baseline analytical method to prove its effectiveness.
Both the \StressLFEM dataset and the baseline implementation will be publicly available.

\end{abstract}

\section{Introduction}
\label{section:intro}

\subsection{Background}

\begin{figure}
\begin{subfigure}{.49\textwidth}
\centering
\includegraphics[scale=0.34]{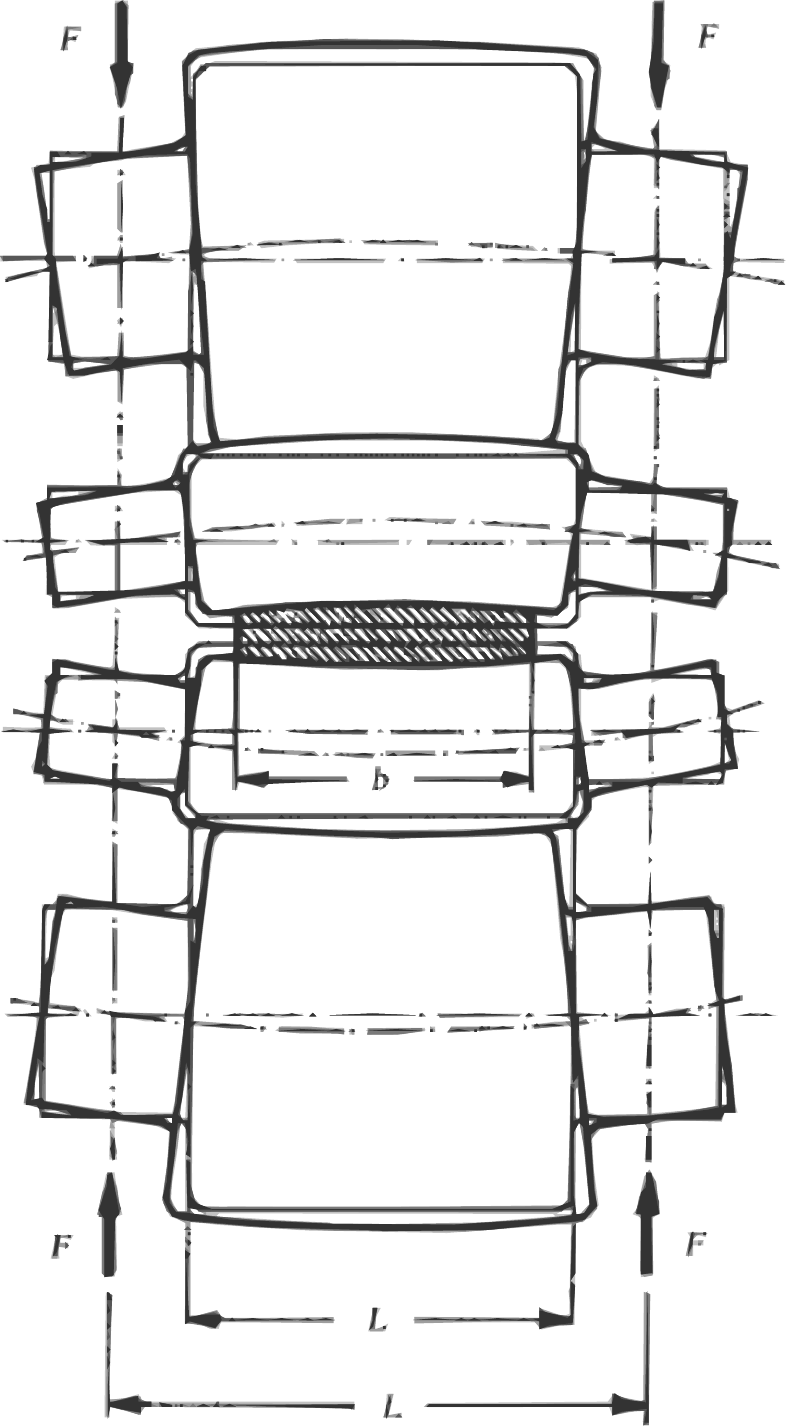}
\caption{}
\label{fig:y_equals_x}
\end{subfigure}
\begin{subfigure}{.49\textwidth}
\centering
\includegraphics[scale=0.5]{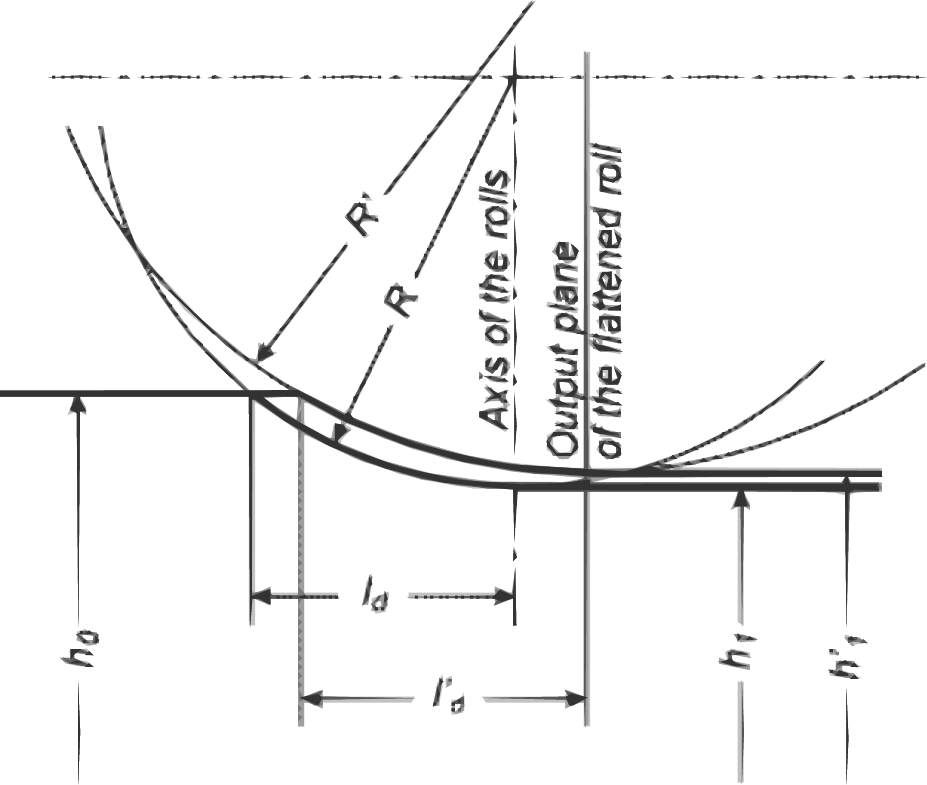}
\caption{}
\label{fig:three_sin_x}
\end{subfigure}
\caption{(a) Shape of the roll gap in the four-high mill system during loading; bold line, the rolls during pass; thin line, the rolls unloaded. (b) Elastic flattening of the rolls: $R^\prime$, radius of the flattened roll; R, radius of the rigid roll; $h_0$, initial thickness; $h_1$, output thickness for the rigid roll;
$h_1$, output thickness for the flattened roll. Images taken from \cite{KNAPINSKI2006257, rooooollling}.}
\label{fig:roll_gap}
\end{figure}

Strain-Induced Martensite transformation in Austenitic Stainless Steels contributes to the improvement of the mechanical properties \cite{seetharaman1981influence}.
Rolling is a primary metalworking process. It is estimated that 80\% of all usable steel stock is processed at least once in a roll mill \cite{shivpuri1989comparative}. The deformation process occurs in a two-roll gap. Forces which are acting within the roll gap, caused by the resistance of the metal against deformation, result in the elastic deformation of the roll, which changes the roll dimensions. The total pressure force, understood as a continuous load acting on the roll faces, causes (a) elastic deflection of the rolls (see \cref{fig:y_equals_x}), and (b) elastic flattening of the rolls, contributing to an increase in the roll gap length and a change in the shape of the contact surface under load (see \cref{fig:three_sin_x}) \cite{KNAPINSKI2006257}. Consequently, a strip with an incorrect cross-sectional contour and diverse dimensional deviations is obtained \cite{KNAPINSKI2006257}. Depending on the type of roll assembly and the design of the rolling stand, different roll gap deformations are observed with the same widths and rolling reductions\cite{yu2008numerical}. 

Flow stress during deformation depends mainly on the strain, strain rate, and temperature, and shows a complex nonlinear relationship with them. Semi-empirical models have been reported to predict the flow stress during deformation \cite{GUPTA2012589}. Recently, with the rise of ML techniques, Artificial Neural Networks (ANN), due to their good generalization performance without needing explicit mathematical and physical knowledge of deformation mechanism, have gained a lot of attention.

\subsection{Literature Review}

Several researchers have studied the behavior of the ASS under tension test in order to investigate the effect of temperature and strain rate on its mechanical properties \cite{peng2004effect,samuel2006dynamic,armas1988strain,jiang2009influence}. The results of the X-Ray pattern diffraction of the deformed ASS 316L show three phases including $\mathcal{E}$-Martensite, Austenite, and $\alpha^\prime$-Martensite \cite{seetharaman1981influence}. The presence of the $\mathcal{E}$-Martensite can increase the site for $\alpha^\prime$-Martensite nucleation \cite{hedayati2010effect}.

Gupta et al. \cite{GUPTA2012589}, predicted the flow stress of ASS 316L, using ANNs with regard to the dynamic strain aging, which occurs in certain deformation conditions.
Knapinski \cite{KNAPINSKI2006257} carried out a numerical simulation of the elastic deflection and flattening of rolls during the rolling process using a commercial computer program designed for simulations of plastic working processes\footnote[2]{FORGE: \url{https://www.transvalor.com/en/forge}.} \cite{dyja1998asymetryczne}, which does not enable the numerical analysis of the state of strain of more than one object at a time. Therefore, the simulation of the elastic deformation of rolls was performed in two steps for each of the strip widths analyzed. 
Moreover, Yu et al. \cite{yu2008numerical} introduced a new type of numerical analysis method: the Contact Element Method (CEM) with two relative coordinates to predict the deflection of work-roll.

Furthermore, the inadequacy of the interface pressure measurement techniques and the ease of measuring roll forces and torques on an actual production rolling mill has led to the development of theorems that predict roll separating forces and the torque \cite{shivpuri1989comparative}. Von Karman \cite{von1925beitrag} was the first to apply the slab method to determine roll pressure and forces in rolling. Later on, Orowan \cite{OROWANTHEORY} developed a comprehensive theorem based on an extension of the slab method by introducing non-homogeneity of plastic deformation of the sheet and elastic deformation of the work roll.

\subsection{Paper Structure and Contributions}
In this study, we follow Orowan's theorem \cite{OROWANTHEORY} and solve the equilibrium equation using an FD approach for both the entry and exit sides of the rolling strip to estimate the mean pressure value which affects the work-roll as a distributed load. The intersection point of the curves of the two sides is called the Neutral Point. 
Secondly, we use the mean pressure utilizing numerical FEA for one-dimensional (1D) cantilevered beam to obtain the roll deflection.
As mentioned above, the best way of finding accurate flow stress of the strip which considers the hardening of the ASS 316L during the rolling is to take advantage of the power of ML methods.
Having the fact that, in this study, we consider multi-pass rolling cases, by increasing the number of passes, the volume fraction of the SIM increases, which leads to an increase in the flow stress and, consequently, increasing the mean pressure and deflection.

Our paper is organized as follows.
In \cref{section:exp}, we present our utilized material and the experimental conditions.
\cref{sec:met} clarifies the proposed numerical method.
Our acquired \StressLFEM dataset and the ANN regression model are presented in \cref{subsection:artificia-neural}.
\cref{section:evaluation} discusses the evaluation of the proposed model. 
Finally, \cref{section:con} states some conclusions and potential future work.
Our obtained \StressLFEM dataset will be publicly available for further investigation of the research community\footnote[3]{The \StressLFEM will be released publicly under this link: \url{https://www.kaggle.com/mahshadlotfinia/Stress316L/}.}.
To encourage reproducibility, the full source code accompanying the document will be publicly accessible as well\footnote{Accessible under this link: \url{https://github.com/mahshadlotfinia/Stress316L/}.}.
\section{Experimental Conditions}
\label{section:exp}

The results of the Quantometer test\footnote{A spectroscopic instrument for measuring the percentage of different metals present in a sample.} (see \cref{tab:chemical-composition}) show the chemical composition of the ASS 316L which is used in this work. In order to consider the hardening as a result of the SIM in specimens, ASS 316L sheets of 4 mm thickness are annealed in the 1030\textdegree{}C for 30 minutes, before deformation. Consequently, specimens will be fully austenitic.

\begin{table}
\centering
\caption{Wt\% of the chemical composition of the ASS 316L.}
\label{tab:chemical-composition}
\begin{tabular}{>{\centering\arraybackslash}m{1cm}>{\centering\arraybackslash}m{1cm}>{\centering\arraybackslash}m{1cm}>{\centering\arraybackslash}m{1cm}>{\centering\arraybackslash}m{1cm}>{\centering\arraybackslash}m{1cm}>{\centering\arraybackslash}m{1cm}>{\centering\arraybackslash}m{1cm}>{\centering\arraybackslash}m{1cm}}
\toprule
\textbf{Fe} & \textbf{Cr} & \textbf{Ni} & \textbf{Mo} & \textbf{Mn} & \textbf{Cu} & \textbf{Si} & \textbf{Co} & \textbf{V}\\
\midrule
Base  & 17.2 &   10.26   &  2.07 & 1.19 & 0.371 & 0.346 & 0.145 & 0.1\\
\bottomrule
\end{tabular}
\end{table}

Before the rolling, we determine the initial dimensions of specimens. Afterwards, with an angular speed of 30 RPM, we roll it in the room temperature, and immediately, measure the final dimensions of the strip.
Finally, we let the strip cool down in the room temperature. For rolling the specimens for more than one pass, we follow the same steps as for the 1-pass rolling condition. In this study, we consider 1 to 7-pass rolling processes. A cold work-roll tool, \ie SLD10\footnote{Cold work die steel with extremely high hardness and excellent toughness.}, in the laboratory scale, is used with the young module of 211 GPa and radius of 75 mm. 
\section{Methodology}
\label{sec:met}

Our proposed numerical method consists of four steps. Firstly, in order to collect the required data using ASS 316L tensile tests, we conduct a series of mechanical tensile tests in a constant room temperature for four different strain rates of $0.001S^{-1}$, $0.00052S^{-1}$, $0.0052S^{-1}$, and $0.052S^{-1}$. 
Also, samples have the ASTME8\footnote{Standard Test Methods for Tension Testing of Metallic Materials.} standard dimensions. 
We should note that all the specimens have been annealed at $1030^\circ$C for 30 minutes. 

In the second step, using our obtained dataset, we train a regression neural network in order to predict the dynamic flow stress during the deformation more accurately.

Third, our main purpose is to estimate the mean pressure that a strip enforces to the roll. 
Therefore, considering the Forward Difference method for the both sides, the yield criterion (\cf\cref{eq:criterioon}), steps of 0.01$^{\circ}$, and the dynamic flow stress from the ANN predictions, the mean pressure as well as the neutral point location, in the Angle-Pressure diagram, will be obtained.

Finally, after calculation of the roll deflection using the obtained mean pressure, we compare the results of the numerical method with those of an analytical method. 
Moreover, we examine the effect of changing the roll radius, to confirm the effectiveness and the accuracy of the proposed model.

\subsection{Equation of Equilibrium}
\label{subsection:equilibrium}

\begin{figure}
\centering
\includegraphics[scale=0.2]{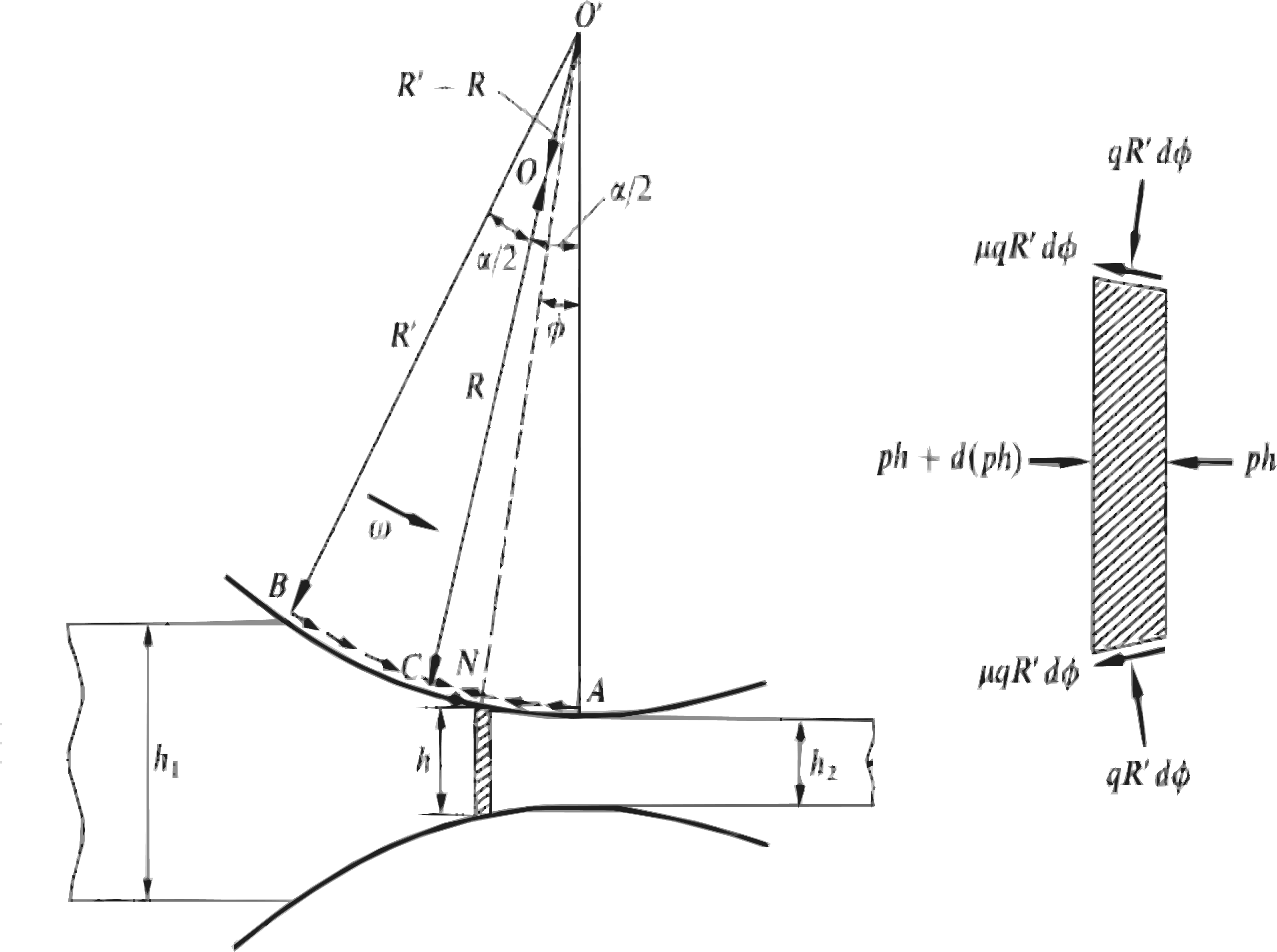}
\caption{Geometry of strip rolling, showing the forces acting on a slice considered on the exit side. Image taken from \cite{chakrabarty1988theory}.}
\label{fig:geometry-ofrolling}
\end{figure}

In Von Karman’s\cite{von1925beitrag} theorem of rolling, it is assumed that at the outset, each element of the strip is uniformly compressed between the rolls while passing through the roll gap. \cref{fig:geometry-ofrolling} shows the geometry of strip rolling, where $R$ and $R^\prime$ represent the radius of the rigid roll and the radius of the flattened roll, respectively.

Firstly, we consider the following yield criterion,

\begin{equation}
    q-p =2k
\label{eq:criterioon},
\end{equation}
where, $k$ represents the shear yield stress. The vertical compression ($q$) of the strip is accompanied by a horizontal force ($p$), which is increasingly compressive as the neutral point is approached from either sides (see \cref{fig:geometry-ofrolling}). Subsequently, the equation of equilibrium can be written as in \cref{eq:quilibri},

\begin{equation}
    \frac{d}{d\phi}(hp)=2qR^{\prime}(\sin\phi\mp \mu\cos\phi)
\label{eq:quilibri},
\end{equation}
where, $h$ represents the thickness of the strip in every related angle ($\phi$) of the contact and  $\mu$ is the friction coefficient between the roll and the strip.
Note that, $k$, generally, varies along the arc of contact. The value of $2k$ at a generic point on the arc of contact is approximately obtained under plane-strain condition, corresponding to an abscissa equal to $\ln(\sfrac{h_1}{h})$. 
Alternatively, the variation of the resulting stress can be estimated by rolling a length of the strip in a succession of passes and carrying out a tensile test at the end of each pass. This gives the tensile yield stress $\sqrt{3}k$ as a function of the thickness ratio ($\sfrac{h_1}{h}$). Considering that the angle of the contact is small, it leads to the governing equation,

\begin{equation}
    \frac{d}{d\phi}(q -2k) \mp 2\mu R^{\prime}q = 4kR^{\prime}\phi
\label{eq:governing_equ},
\end{equation}
where, the value of $\phi$ can be calculated using the following equation,

\begin{equation}
    h = h_2 + 2R^{\prime}(1-\cos\phi) \approx h_2 + R^{\prime}\phi^2
\label{eq:equ1_4}.
\end{equation}
In \cref{eq:equ1_4}, $h_1$ and $h_2$ show the initial and final thicknesses of the rolled plate, respectively.

\subsection{Finite Difference (FD) Formulation}
\label{subsection:finite-formulation}

To determine the distributed load applied to the roll, we consider the FD approach for the investigation of the equilibrium equation (\cf\cref{eq:quilibri}). For both the entry and exit sides, we consider the Forward Difference scheme.

As long as the exit side is concerned, we use the following (\cf\cref{eq:equ1_5}) Forward Difference formulation of \cref{eq:governing_equ}. Applying the initial and final thicknesses of the strip on \cref{eq:equ1_4}, to estimate $\phi_1$, and assuming that $p=0$ (consequently, $2k=q_0$), we obtain the initially needed factors for the first step of \cref{eq:equ1_5}. In both the entry and exit sides, $i$ represents the index of the steps. 

\begin{equation}
    \frac{h_i (q_i - 2k_{i+1} - q_i + 2k_i)}{\phi_{i+1} - \phi_i} + 2\mu Rq_0 = 4kR\phi_i
\label{eq:equ1_5}
\end{equation}
Similar to the exist side, the Forward Difference approach is used for the entry side, as explained in \cref{eq:entryside},

\begin{equation}
    \frac{h_i (q_i - 2k_{i+1})}{\phi_{i+1}} - 2\mu Rq_0 = 0
\label{eq:entryside}.
\end{equation}

Moreover, when the calculated pressures in the entry and exit sides (\cf \cref{eq:equ1_5} and \cref{eq:entryside}) are equal, we could assume it as a neutral pressure point in the corresponding angle. Consequently, the mean pressure is calculated according to \cref{eq:mean_pressure}.

\begin{equation}
    \bar{P} = \frac{\int_{\alpha}^{0} q \,d\phi}{\Delta \phi}
\label{eq:mean_pressure}
\end{equation}

\subsection{Deflection and Work-Roll Characteristics}
\label{deflectionnnn}

\cref{eq:galerkinn} shows the governing equation for the deflection of the beam. Here, $W$ represents the deflection in the $y$ direction, while, $EI$ is the product of the modulus of elasticity $E$ and the moment of inertia $I$ (see \cref{eq:equu1_9}).

\begin{equation}
    \frac{d^2}{dx^2} \left( EI\frac{d^2W}{dx^2} \right) = p(x)
\label{eq:galerkinn},
\end{equation}
where,

\begin{equation}
    I = \frac{\pi d^4}{64}
\label{eq:equu1_9}.
\end{equation}

We take the Galerkin Weighted-Residual approach into consideration to obtain the Finite Element formulation for the beam element. Every element has two nodes. However, four requirements must be met: the deflections and slopes at the two nodes. At the boundary of each element, both the deflection and the slope are assumed to be continuous, \ie the parameter function is $C^I$-continuous rigid \cite{stasa1985applied}. 

Throughout the last step, we consider the roll to be a 1D cantilevered beam with a circular cross-section, and we investigate its deflection during the deformation. According to real experimental conditions, while the total length of the beam is 20 cm, only 22.5\% of it is under the pressure of the distributed load. Because of the plane-strain condition, the width of the strip does not change during the multi-pass rolling.


\section{Artificial Neural Network (ANN)}
\label{subsection:artificia-neural}

Unlike the other groups of metals, ASS 316L has an unpredictable Strain-Stress curve. Thus, we conducted a series of tensile tests at different strain rates.
In any stage of deformation, due to the work hardening and also the SIM, the flow stress dynamically changes. Furthermore, our collected dataset is obtained using experimental tests which are subject to error and would lead to deviations in our Strain-Stress model.
Therefore, we take advantage of the ML techniques and utilize a regression analysis model in order to calculate the flow stress more accurately. In the following, we first go through our obtained dataset in order to give an overview of our goal.

\subsection{\StressLFEM: Dataset of Strain \& Stress of Stainless Steel 316L During Cold Tension}

As mentioned above, our collected dataset is subject to errors. Therefore, we aim at benefiting from ML methods and consequently, following the idea in \cite{RETWEET}, we convert our obtained data into an ML dataset, called \StressLFEM. To the best of our knowledge, this is the first public dataset of Strain-Stress values for the ASS 316L druing rolling.

We conducted four sets of Uniaxial Tensile Tests in $0.001S^{-1}$, $0.00052S^{-1}$, $0.0052S^{-1}$, and $0.052S^{-1}$ strain rates in the room temperature with ASS 316L samples.
According to the ASTME8 standard, the ASS 316L sheets with initial thickness of 4 mm, width of 6 mm, and Gage length of 32 mm were utilized for the tensile tests using a compression test machine\footnote{Electro Mechanic Instron 4208.}. 
The output results are in the forms of extension (in mm) and force (in N), which were converted to the true-strain and true-stress values, respectively. The data conversion procedure was done by considering the cross-section of the loaded force, which for our case was $24$ mm$^2$.
Finally, we ended up with 15,858 different Strain-Stress values at four different strain rates.

\subsection{ANN Model}
\label{sub:archite}

\begin{figure}
\begin{subfigure}{.49\textwidth}
\centering4\includegraphics[scale=0.5]{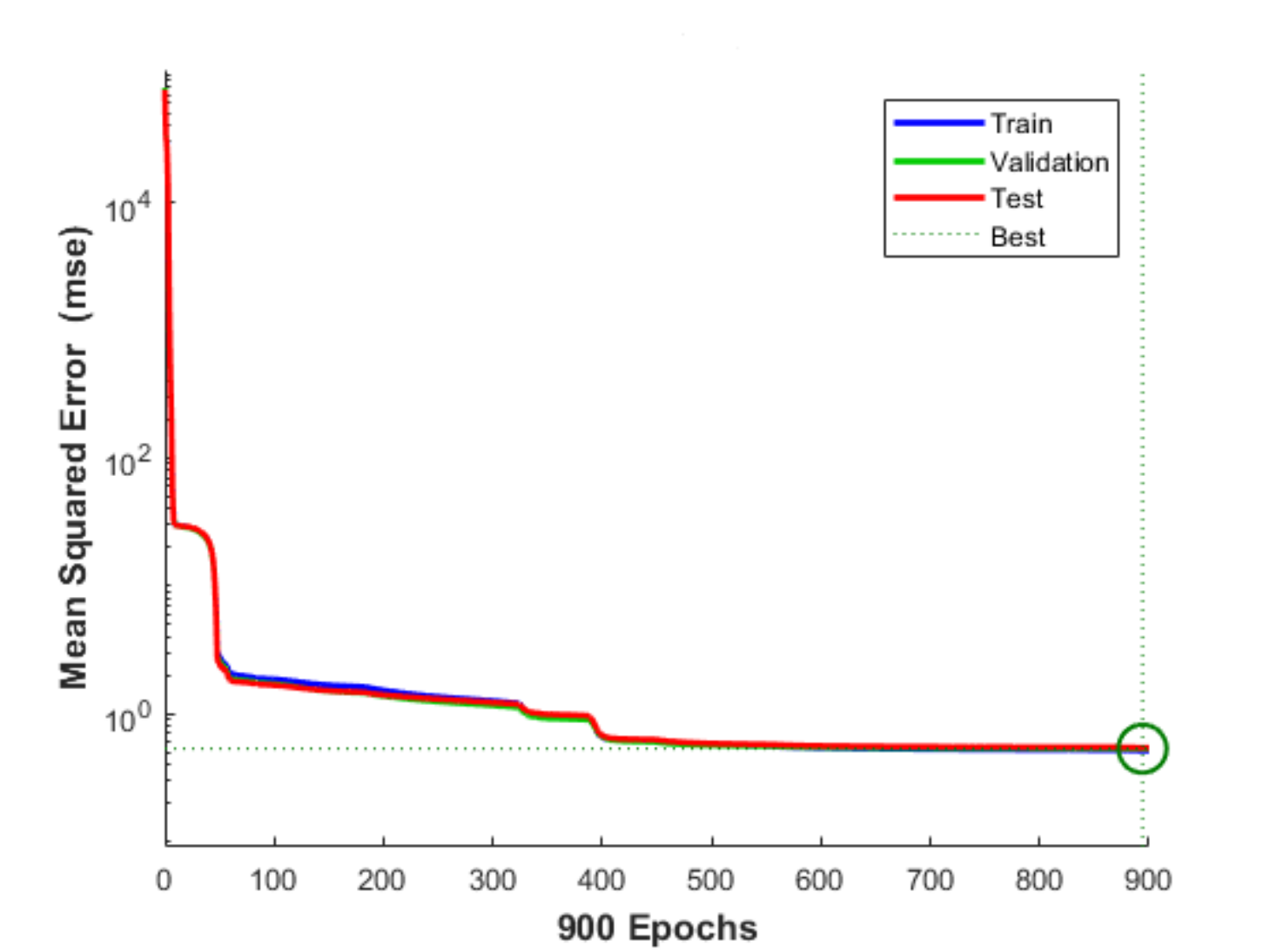}
\caption{}
\label{fig:training_curves}
\end{subfigure}
\begin{subfigure}{.49\textwidth}
\centering
\begin{tikzpicture}
\begin{axis}[
                 grid=both,
                 grid style={solid,gray!30!white},
                 xlabel={Strain},
                 ylabel={Stress},
                 xmin=0,
                ]
\addplot[blue, only marks, mark size=.9pt] table [x=strain, y=stress, col sep=comma] {figs/x_y_initial.csv};
\addplot[red, only marks, mark size=.9pt] table [x=strain, y=stress, col sep=comma] {figs/x_y_final.csv};
\legend{Original data, Predicted data}
\end{axis}
\end{tikzpicture}
\caption{}
\label{fig:stress-strain_initial_final}
\end{subfigure}
\caption{(a) Loss curves for training, validation, and the test data. (b) Strain-Stress diagram of the data before and after using the neural network. Note that, we have down-sampled the original and predicted data in (b) with a factor of 50 in order to be able to visualize the improvement resulted from using the network.}
\label{fig:stress-strain_training}
\end{figure}
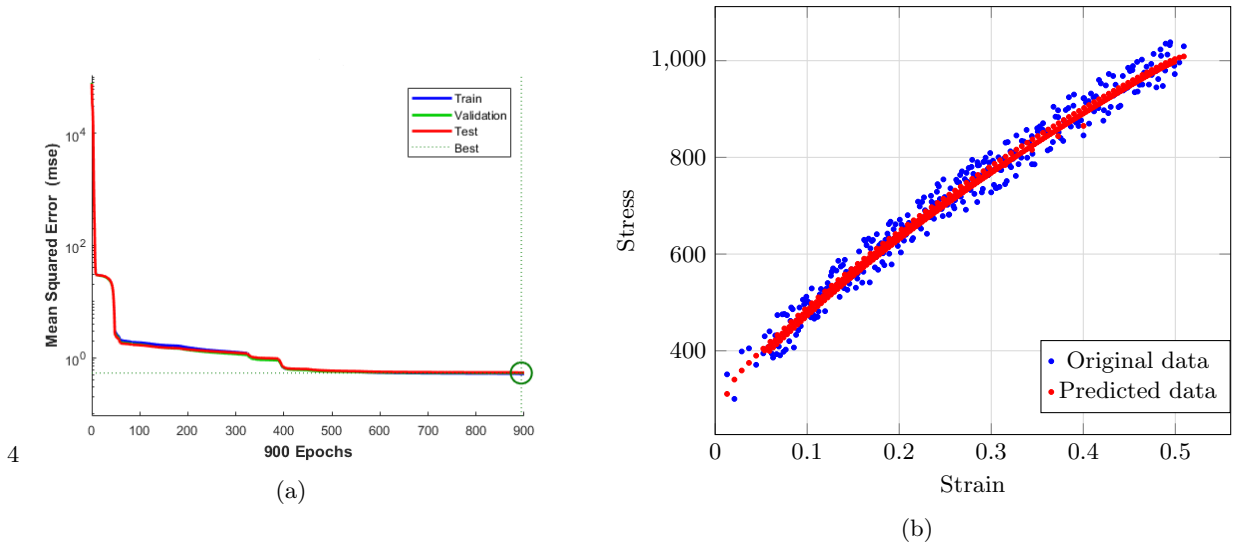

Looking at the general trend of our dataset (see \cref{fig:stress-strain_initial_final}), we observe that the best way to go is simply linear regression (\cf\cref{eq:regression}).

\begin{equation}
    \mathbf{y} = \mathbf{X}\bm{\beta} + \bm{\epsilon}, \,\,\,\,\,\,\,\,\, \text{where,} \,\,\,\,\,\,\,\,\, 
    \mathbf{X} =
\begin{bmatrix}
1 & x_{11} & x_{12} \\
1 & x_{21} & x_{22} \\
\vdots & \vdots & \vdots \\
1 & x_{n1} & x_{n2}
\end{bmatrix},
\label{eq:regression}
\end{equation}
and $\mathbf{y}$ corresponds to the output prediction, which is stress in our case and the rows of $\mathbf{X}$ correspond to the feature vectors of our data, which correspond to the strain and strain rate values in our case.
Furthermore, $\bm{\beta}$ and $\bm{\epsilon}$ represent the weights and the biases of the network, respectively, while, $n$ is the total number of the data.

\begin{figure}
\centering
\includegraphics[scale=0.5]{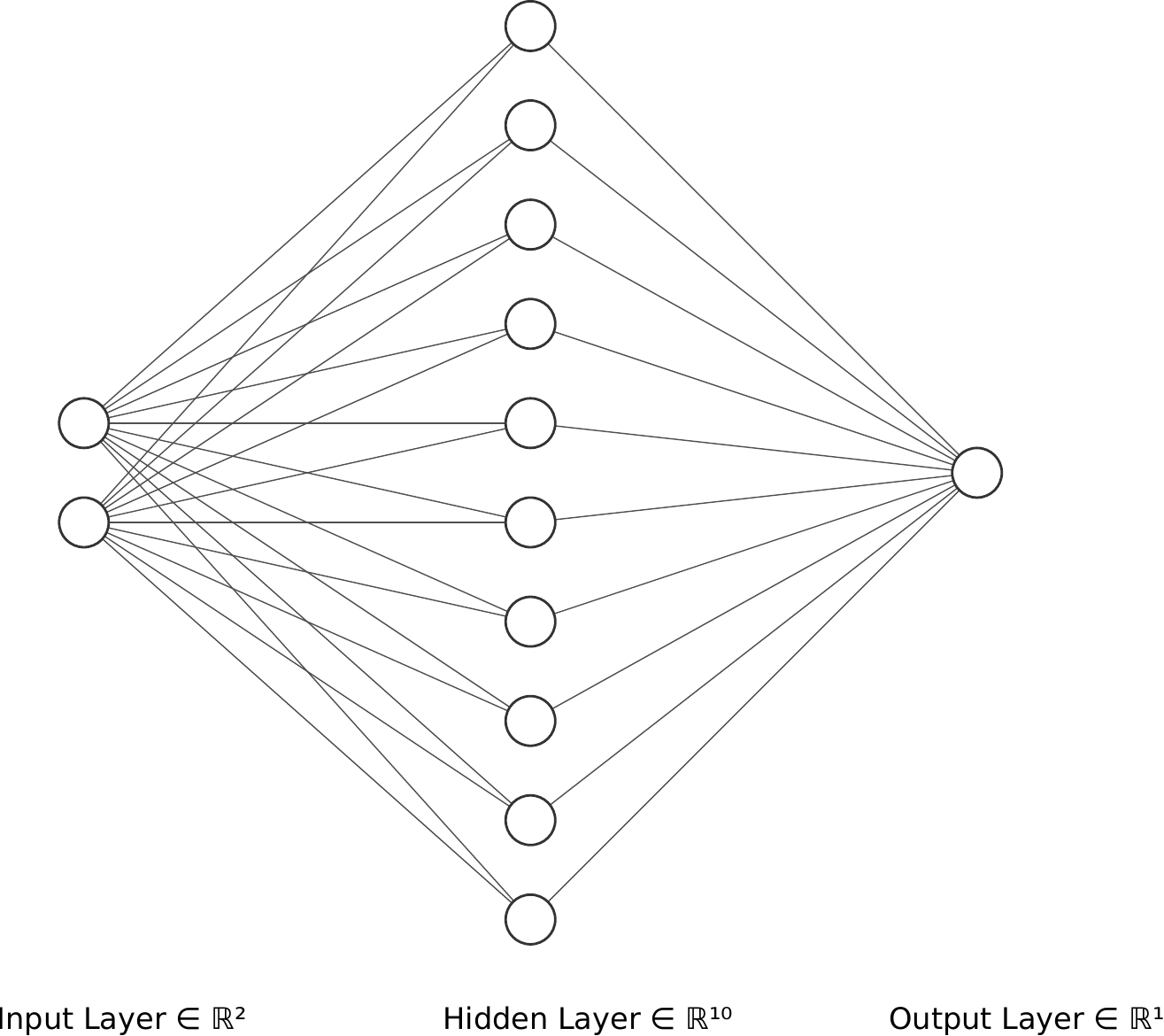}
\centering
\caption{Model architecture.}
\label{fig:architecture}
\end{figure}

Additionally, due to the fact that our \StressLFEM dataset contains only data with two-dimensional feature vectors, we do not need to go for Deep Learning (DL) methods.
ANNs are particularly good at fitting functions and with solely a simple neural network, we can already fit practical functions. Therefore, we propose a simple shallow feed-forward fully-connected network, which only contains two layers. \cref{fig:architecture} illustrates our proposed linear regression network which has only a single hidden layer, and the Sigmoid function is selected as the activation function.
Furthermore,
Mean Squared Error (MSE) (\cf\cref{eq:mse}) is chosen as our loss function and the Levenberg-Marquadt Backpropagation method is used as the optimizer to solve the gradient-descent problem.

\begin{equation}
    MSE = \frac{\sum_{i=1}^{n} (y_i - y^\prime_i)^2}{n}
\label{eq:mse},
\end{equation}
where, $y_i$ and $y^\prime_i$ represent the actual and the predicted stress values, respectively.
We use 70\% of the dataset as training data, 15\% as validation, and the rest as the test data and train the network for 900 epochs. \cref{fig:training_curves} illustrates the training, validation, and the test curves. We observe that using only the proposed simple architecture, the validation and test loss values are already good and we do not need to go for deeper layers. Moreover, \cref{fig:stress-strain_initial_final} shows the scatter Strain-Stress diagrams of the data, before and after using the network in which, we easily observe the improvement in accuracy.

\section{Evaluation and Results}
\label{section:evaluation}

Our proposed model (\cf\cref{sec:met}) is, undoubtedly, an effective and beneficial model to obtain an accurate value of the roll deflection during the rolling of the ASS 316L sheets. Also, we could utilize this model for other types of ASS with the same characteristics. However, the corresponding dataset for training the ANN must be provided for any type of steel.
Furthermore, when specimen rolls for one pass, its thickness varies from 4 to 3.24 mm. Through the width, no variation is observed, which shows that the plane-strain condition prevents the plate to be altered through its width. The same happens for more than 1-pass.
\cref{tab:multi-pass-thickness} shows the final thicknesses of specimens which are measured using digital calipers.
It can be seen that as the number of passes increases, the reduction in thickness decreases. This phenomenon is a result of the SIM formation. Also, as the amount of strain increases, the volume fraction of the SIM increases as well, and consequently, the hardness of the plates increases. This makes the specimens to be deformed less and consequently, have less thickness reductions by increasing the number of passes.

\begin{table}[t]
\centering
\caption{Final thicknesses of specimens after multi-pass rolling in the room temperature, given in mm.}
\label{tab:multi-pass-thickness}
\begin{tabular}{>{\centering\arraybackslash}m{1.3cm}>{\centering\arraybackslash}m{1.3cm}>{\centering\arraybackslash}m{1.3cm}>{\centering\arraybackslash}m{1.3cm}>{\centering\arraybackslash}m{1.3cm}>{\centering\arraybackslash}m{1.3cm}>{\centering\arraybackslash}m{1.3cm}>{\centering\arraybackslash}m{1.3cm}>{\centering\arraybackslash}m{1.3cm}}
\toprule
\textbf{2-Pass} & \textbf{3-Pass} & \textbf{4-Pass} & \textbf{5-Pass} & \textbf{6-Pass} & \textbf{7-Pass} \\
\midrule
2.8  &  2.48  &  2.23   &  1.99  &  1.74  & 1.44 \\
\bottomrule
\end{tabular}
\end{table}

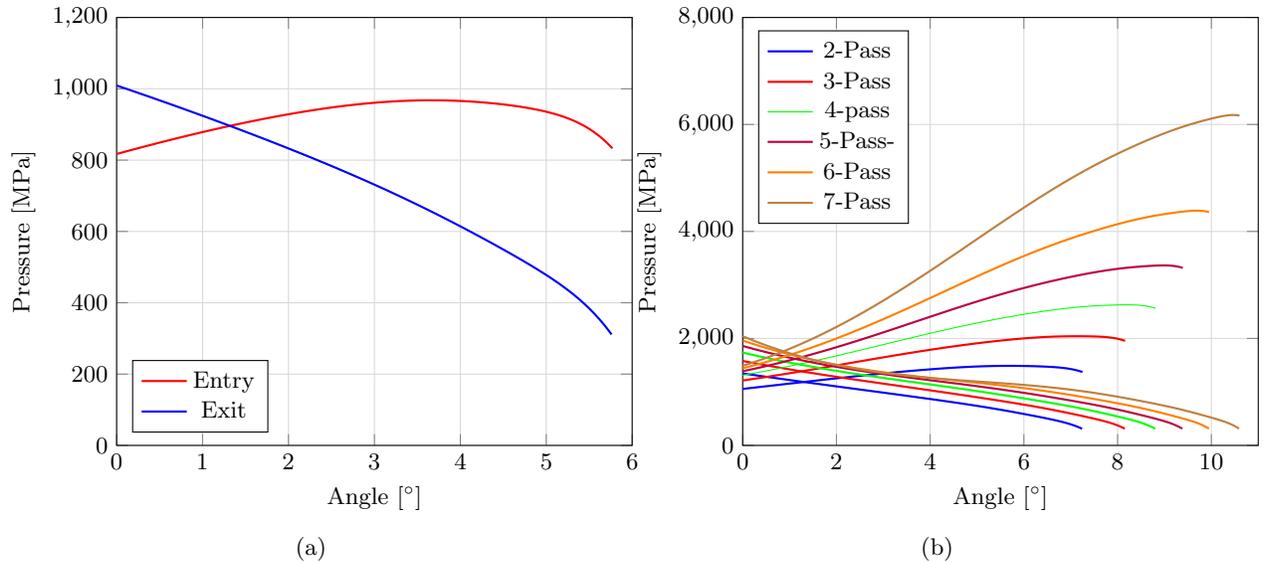
\begin{figure}
\begin{subfigure}{.49\textwidth}
\begin{tikzpicture}
\begin{axis}[
                 grid=both,
                 grid style={solid,gray!30!white},
                 xlabel={Angle [$^{\circ}$]},
                 ylabel={Pressure [MPa]},
                 legend pos=south west,
                 xmin=0,
                 xmax=6,
                 ymin=0,
                 ymax=1200,
                ]
\addplot[thick, red] table [x=Step, y=Value, col sep=comma] {figs/Book2.csv};
\addplot[thick, blue] table [x=Step, y=Value, col sep=comma] {figs/Book3.csv};
\legend{Entry, Exit}
\end{axis}
\end{tikzpicture}
\caption{}
\label{entry-exit}
\end{subfigure}
\begin{subfigure}{.49\textwidth}
\centering
\begin{tikzpicture}
\begin{axis}[
                 grid=both,
                 grid style={solid,gray!30!white},
                 xlabel={Angle [$^{\circ}$]},
                 ylabel={Pressure [MPa]},
                  legend pos=north west,
                 xmin=0,
                 xmax=11,
                 ymin=0,
                 ymax=8000,
                ]
\addplot[thick, blue] table [x=Step, y=Value, col sep=comma] {figs/2pasenter.csv};
\addplot[thick, blue] table [x=Step, y=Value, col sep=comma] {figs/2pasexit.csv};
\addplot[thick, red] table [x=Step, y=Value, col sep=comma] {figs/3pasenter.csv};
\addplot[thick, red] table [x=Step, y=Value, col sep=comma] {figs/3pasexit.csv};
\addplot[green] table [x=Step, y=Value, col sep=comma] {figs/4pasenter.csv};
\addplot[thick, green] table [x=Step, y=Value, col sep=comma] {figs/4pasexit.csv};
\addplot[thick, purple] table [x=Step, y=Value, col sep=comma] {figs/5pasenter.csv};
\addplot[thick, purple] table [x=Step, y=Value, col sep=comma] {figs/5pasexit.csv};
\addplot[thick, orange] table [x=Step, y=Value, col sep=comma] {figs/6pasenter.csv};
\addplot[thick, orange] table [x=Step, y=Value, col sep=comma] {figs/6pasexit.csv};
\addplot[thick, brown] table [x=Step, y=Value, col sep=comma] {figs/7pasenter.csv};
\addplot[thick, brown] table [x=Step, y=Value, col sep=comma] {figs/7pasexit.csv};
\legend{2-Pass,,3-Pass,,4-pass,,5-Pass-,,6-Pass,,7-Pass,,}
\end{axis}
\end{tikzpicture}
\caption{}
\label{fig:angle-compare}
\end{subfigure}
\caption{(a) Intersection of the Angle-Pressure diagrams of the entry and exit sides for the 1-pass rolled strip, which represents the neutral point. (b) Angle-Pressure diagrams of the multi-pass rolled ASS 316L strips in the room temperature for the entry (upper curves) and exit (lower curves) sides and for the 2 to 7-pass rolled strips, which represent the neutral points of the 2 to 7-pass rolled strips, respectively. We observe that, increasing pass numbers, the neutral point location remains almost the same while, the maximum pressure increases.}
\label{fig:three_graphs}
\end{figure}

\cref{entry-exit} shows the intersection point of the entry (\cf\cref{eq:entryside}) and exit (\cf\cref{eq:equ1_5}) sides in the Angle-Pressure diagrams.
The angle of the neutral point for 1-pass rolling is 1.31$^{\circ}$ at where, the strip enforces the maximum pressure on the tolls. By comparing the angle of the neutral points for the next 6 passes of rolling, which can be seen in \cref{fig:angle-compare}, we conclude that the angle is almost constant, rather the total pressures in every pass increase.

As our main evaluation metric for the proposed method (\cf\cref{sec:met}), \cref{eq:first-validity} is used for determining the neutral point of rolling, in an analytical manner, instead of the mentioned numerical analysis \cite{wusatowski1969fundamentals}. 

\begin{equation}
    \sin(\gamma_n) = \frac{\sin \alpha}{2} + \frac{\cos \alpha -1}{2\mu}
\label{eq:first-validity}.
\end{equation}
The contact angle of tools ($\gamma$), using \cref{eq:equ1_4}, is equal to 5.77$^{\circ}$ and, the analytical value of the neutral point is 1.42$^{\circ}$. Furthermore, the maximum value of the pressure using analytical analysis is 876 MPa. Whereas, in the case of using numerical analysis, it changes to 896 Mpa at 1.31$^{\circ}$.
Consequently, this FD model is a reliable method to predict the mean pressure value more accurately.

The mean pressure, which the material enforces to the work-roll tools and could be obtained using \cref{eq:mean_pressure} for the 1-pass roll, is 121 MPa. In the next step of the investigation, we should input the mean pressure and the number of elements to the FEM model.
According to the dimensions of the tools and plate, the optimal number of utilized elements is 100. The values of the mean pressure for 2 to 7-pass rolls are represented in \cref{tab:multi-pass-pressure}. The highest value of deflection can be seen in the middle of the work-roll tool. This is due to the design of the tool which causes the strip to go through the center of the work-roll tools, during the rolling process. 

\begin{table}
\centering
\caption{Mean pressure values as distributed loads, which cause deflection in the rolls, given in MPa. By increasing the thickness reduction by increasing the number of passes, the roll deflection increases.}
\label{tab:multi-pass-pressure}
\begin{tabular}{>{\centering\arraybackslash}m{1.3cm}>{\centering\arraybackslash}m{1.3cm}>{\centering\arraybackslash}m{1.3cm}>{\centering\arraybackslash}m{1.3cm}>{\centering\arraybackslash}m{1.3cm}>{\centering\arraybackslash}m{1.3cm}>{\centering\arraybackslash}m{1.3cm}>{\centering\arraybackslash}m{1.3cm}>{\centering\arraybackslash}m{1.3cm}}
\toprule
\textbf{2-Pass} & \textbf{3-Pass} & \textbf{4-Pass} & \textbf{5-Pass} & \textbf{6-Pass} & \textbf{7-Pass} \\
\midrule
236 & 335 & 418 & 498 & 575 & 658  \\
\bottomrule
\end{tabular}
\end{table}

\begin{figure}
\centering
\begin{tikzpicture}
\centering
\begin{axis}[
                 grid=both,
                 grid style={solid,gray!30!white},
                 xlabel={$x$ [m]},
                 ylabel={Deflection [mm]},
                  legend pos=north east,
                 xmin=0,
                 xmax=0.2,
                 ymin=0,
                ]
\addplot[thick, black] table [x=Step, y=Value, col sep=comma] {figs/Deflection1pas.csv};
\addplot[thick, blue] table [x=Step, y=Value, col sep=comma] {figs/def2pas.csv};
\addplot[thick, red] table [x=Step, y=Value, col sep=comma] {figs/def3pas.csv};
\addplot[thick, green] table [x=Step, y=Value, col sep=comma] {figs/def4pas.csv};
\addplot[purple] table [x=Step, y=Value, col sep=comma] {figs/def5pas.csv};
\addplot[thick, orange] table [x=Step, y=Value, col sep=comma] {figs/def6pas.csv};
\addplot[thick, brown] table [x=Step, y=Value, col sep=comma] {figs/def7pas.csv};
\legend{1-Pass,2-Pass,3-pass,4-Pass-,5-Pass,6-Pass,7-Pass}
\end{axis}
\end{tikzpicture}
\caption{Comparing deflection values along the roll widths, which are caused by the multi-pass rolling of the ASS 316L in the room temperature. We observe that by increasing the number of passes, the deflection value increases. Moreover, due to the plain-strain condition, the deflection values vary according to the normal distribution along the width of the roll.}
\label{fig:deflection-comapre}
\end{figure}
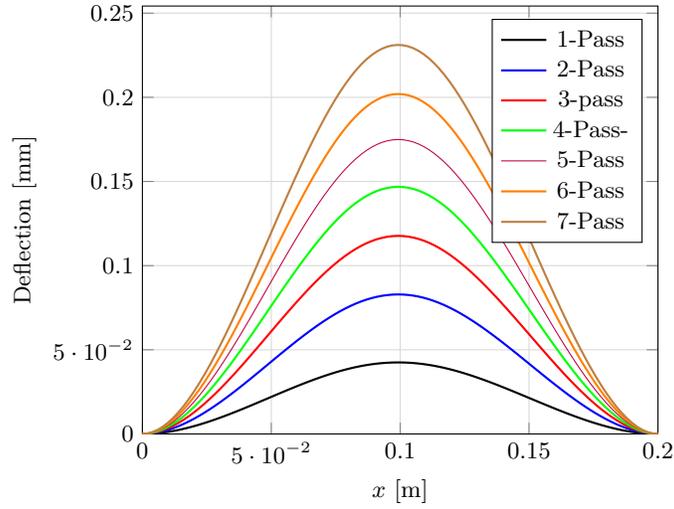

\cref{fig:deflection-comapre} shows by increasing the number of passes, due to the hardening of the strip and thickness reduction, the mean pressure increases and consequently, the deflection increases.
Moreover, due to the plain-strain condition, the deflection values vary according to the normal distribution along the width of the roll ($x$).


Also, we compare the analytical analysis of the deflection at $r=0$, using \cref{eq:deflectionmax}, which represents the maximum deflection of the cantilevered beam, with the numerical results from our model (\cf\cref{sec:met}).
The proposed numerical model (\cf\cref{sec:met}) considers the effectiveness of the rolled sheet material while, the analytical model considers only the roll material effects. Thus, it would be more beneficial to use the numerical approach. 
 \begin{equation}
    \text{Deflection(max)} = \frac{pa^2}{16 \pi D}
\label{eq:deflectionmax},
\end{equation}
where, the material parameter of the rolls ($D$) is calculated as in the following,
\begin{equation}
    D = \frac{ET^3}{12(1 -v)}
\label{eq:equu1_17}.
\end{equation}

\begin{figure}
    \centering
    \includegraphics[scale=.17]{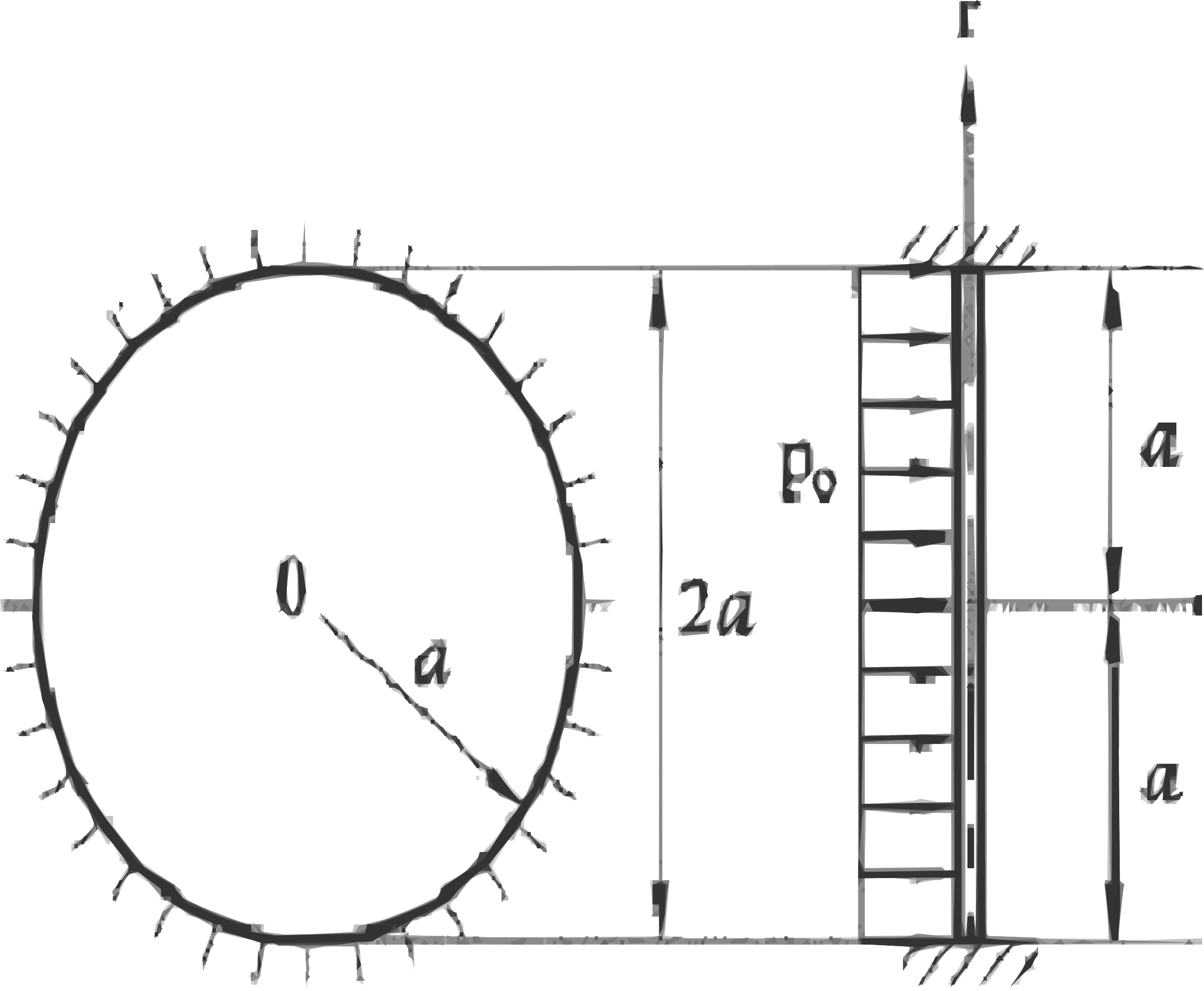}
    \caption{The circular cross-section of the roll, under pressure of the plate during the rolling process. Image taken from \cite{hicks1985standard}.}
    \label{fig:circushape}
\end{figure}

\cref{fig:circushape} illustrates the circular cross-section of the beam which we consider as the cross-section of the roll. $v$ and $E$ are the material characteristics of the rolls, and $t$ represents the thickness of the roll which we assume to be equal to 1.
\begin{figure}
\begin{subfigure}{.49\textwidth}
\begin{tikzpicture}
\centering
\begin{axis}[
                 grid=both,
                 grid style={solid,gray!30!white},
                 xlabel={x [m]},
                 ylabel={Deflection [mm]},
                 legend pos=north east,
                 xmin=0,
                 xmax=0.2,
                 ymin=0,
                 ymax=0.000015,
                ]
\addplot[thick, red] table [x=Step, y=Value, col sep=comma] {figs/75mm.csv};
\addplot[thick, green] table [x=Step, y=Value, col sep=comma] {figs/95mm.csv};
\legend{75mm, 95mm}
\end{axis}
\end{tikzpicture}
\centering
\caption{}
\label{fig:radiuss}
\end{subfigure}
\begin{subfigure}{.49\textwidth}
\begin{tikzpicture}
\centering
\begin{axis}[
                 grid=both,
                 grid style={solid,gray!30!white},
                 xlabel={True-Strain },
                 ylabel={Deflection [mm]},
                  legend pos=south east,
                 xmin=0,
                 xmax=1,
                 ymin=0,
                 ymax=0.3,
                ]
\addplot[thick, blue] table [x=Step, y=Value, col sep=comma] {figs/analytical.csv};
\addplot[thick, orange] table [x=Step, y=Value, col sep=comma] {figs/numerical.csv};
\legend{Analytical Analysis, Numerical Analysis}
\end{axis}
\end{tikzpicture}
\caption{}
\label{fig:num-ana}
\end{subfigure}
\caption{Evaluation results. (a) Deflection values of the roll with different radii of 95 mm and 75 mm. We observe that increasing the radius of the roll leads to a decrease in the deflection value. (b) The deviation between the analytical and numerical analyses of the maximum deflection of the roll for seven different thickness reductions of the strip shows the effectiveness of our proposed numerical method (\cf\cref{sec:met}).}
\end{figure}
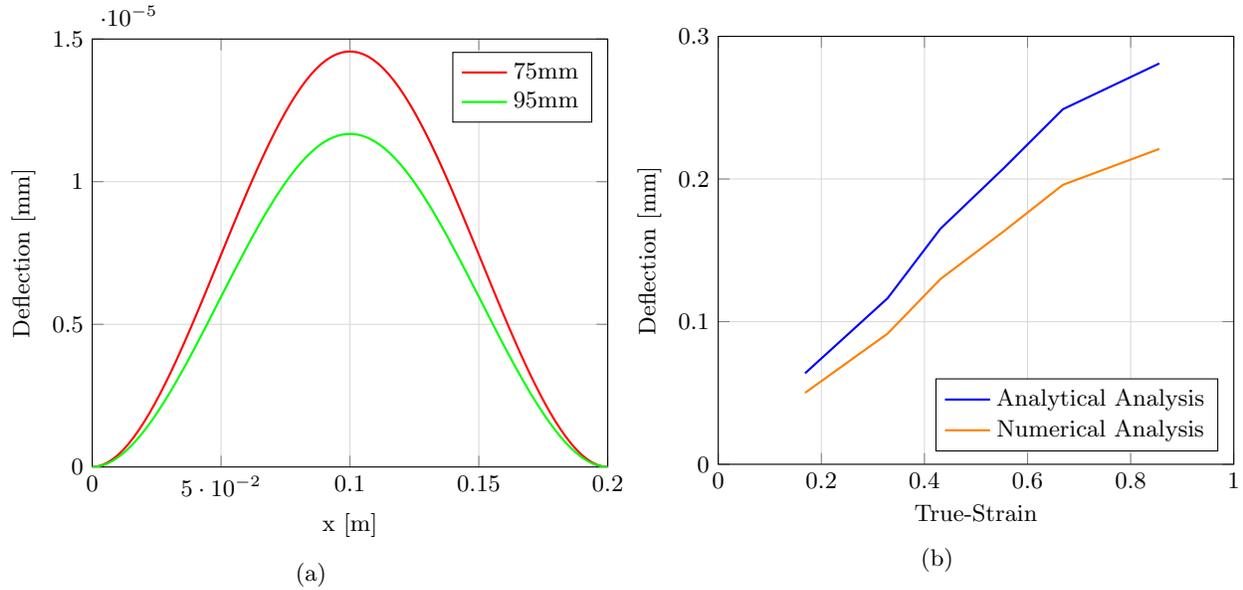
\cref{fig:num-ana} shows significant differences between the analytical and numerical analyses, which proves the accuracy of the proposed numerical method (\cf\cref{sec:met}). In this case, as well as the accuracy of the results, we could obtain the deflection of every element of the work-roll tool. Furthermore, by comparing the two radii of 75 mm and 95 mm (see \cref{fig:radiuss}), we observe that the deflection corresponding to the former is less, which confirms the validity of another condition of our proposed model (\cf\cref{sec:met}).


\section{Conclusion and Discussion}
\label{section:con}

In this work, we proposed a generalized model (\cf\cref{sec:met}) 
to predict more realistically the roll deflection of the Austenitic Stainless Steel 316L.
Instead of using a constant value of flow stress during the multi-pass rolling process, we used a Finite Difference formulation of the equilibrium equation, which led to the estimation of the mean pressure, which the strip enforces to the rolls during deformation. Afterwards, using Finite Element Analysis, we calculated the deflection of the work-roll tools.
Furthermore, we utilized a shallow feed-forward regression neural network for predicting the flow stress values more accurately and accounting for human errors during mechanical tensile tests.
To the best of our knowledge, this is the most generalized model, specific to the ASS 316L, which considers dynamic flow stress and the Strain-Induced Martensite formation during the rolling, using FEA and a Machine Learning approach.
Comparison between the results of our proposed method and a baseline analytical method showed the superiority of our model.
Additionally, we created \StressLFEM, the first public dataset for Strain-Stress values of ASS 316L during cold tension from real experiments.

For future work, we will consider the temperature factor in the tensile tests by conducting different sub-zero temperatures tests in order to extend our \StressLFEM dataset. Moreover, as the other Austenitic Stainless Steels might have the same features as for the 316L, it would be beneficial to generalize our dataset by conducting tensile tests on other steels.
Furthermore, with increasing the dimensionality of the feature vectors, we could utilize deeper networks. Training different networks and ensembling the resulting fits would also be beneficial.
%
%
%

\bibliographystyle{splncs04}
\bibliography{mybib}

\end{document}